\documentclass[conference]{IEEEtran}
\IEEEoverridecommandlockouts

\usepackage[pagebackref,breaklinks,colorlinks,citecolor=green]{hyperref}
 
\usepackage{authblk}
\usepackage{cite}
 \usepackage{multirow}
\usepackage{amsmath,amssymb,amsfonts}
\usepackage{algorithmic}
\usepackage{graphicx}
\usepackage{textcomp}
\usepackage{xcolor}
\usepackage{caption}
\usepackage{bbding}
\usepackage{svg}

\usepackage[ruled,linesnumbered]{algorithm2e}
\def\BibTeX{{\rm B\kern-.05em{\sc i\kern-.025em b}\kern-.08em
    T\kern-.1667em\lower.7ex\hbox{E}\kern-.125emX}}
\begin{document}

\title{Masked Conditional Diffusion Model for Enhancing Deepfake Detection}

\author{
   
    \textbf{Tiewen Chen${^{1}}$}, 
    \textbf{Shanmin Yang$^{1,*}$},
    \textbf{Shu Hu${^2}$},
    \textbf{Zhenghan Fang${^3}$},
    \textbf{Ying Fu${^1}$},
    \textbf{Xi Wu${^1}$},
    \textbf{Xin Wang$^{4,*}$} \thanks{ $^{*}$ Corresponding authors.}
}
\affil{
    
    \textsuperscript{\rm 1} Chengdu University of Information Technology\\
    
    \textsuperscript{\rm 2} Purdue University in Indianapolis \\
    \textsuperscript{\rm 3} Johns Hopkins University \\
    
    \textsuperscript{\rm 4} University at Albany, State University of New York (SUNY)\\
    \textsuperscript{} yangsm@cuit.edu.cn, 
    
    \textsuperscript{} xwang56@albany.edu 
}

\maketitle

\begin{abstract}
Recent studies on deepfake detection have achieved promising results when training and testing faces are from the same dataset. However, their results severely degrade when confronted with forged samples that the model has not yet seen during training. 
In this paper,
deepfake data to help detect deepfakes. this paper present
we put a new insight into diffusion model-based data augmentation, and propose a Masked Conditional Diffusion Model (MCDM) for enhancing deepfake detection.
It generates a variety of forged faces from a masked pristine one, encouraging the deepfake detection model to learn generic and robust representations without overfitting to special artifacts.  
Extensive experiments demonstrate that forgery images generated with our method are of high quality and 
helpful to improve the performance of deepfake detection models. 
 
\end{abstract}


\section{Introduction}

The advent of deep learning-based generative techniques, as exemplified by technologies such as generative adversarial networks (GANs) \cite{goodfellow2014generative} and diffusion models \cite{Nichol_Dhariwal_2021},   
has significantly enhanced the quality of generated images, known as deepfake.
These technologies can manipulate the appearance and voice of real individuals.
Consequently, deepfake applications are capable of producing fabricated videos and audio recordings depicting individuals engaging in or uttering malicious content, thereby posing risks in terms of disinformation, fraud, blackmail, and impersonation.

Furthermore, the utilization of deepfake technology has the potential to erode the credibility of authentic media. This is due to the inherent skepticism that may arise, as individuals may question the authenticity of any evidence or testimony that can be easily manipulated. 
The widespread use of deepfake faces poses a substantial threat to social trust and public safety \cite{Masood_Nawaz_Malik_Javed_Irtaza_Malik_2023}.
Hence, there is an imperative need to develop effective methods for detecting and mitigating the impact of deepfake attacks. Additionally, efforts should be directed toward raising public awareness of the potential risks and ethical implications associated with this rapidly advancing technology.

In recent years, a growing number of researchers have been exploring deepfake detection methods 
to prevent the abuse of fake face images.
However, as it is impossible to collect all kinds of face images generated by all deepfake technologies,
the scarcity of diverse deepfake data becomes a bottleneck to improving the performance of deepfake detection models. 
Though diffusion model \cite{saharia2022palette,dhariwal2021diffusion,ye2023synthetic,Nichol_Dhariwal_2021,Rombach_Blattmann_Lorenz_Esser_Ommer_2022,du2023arsdm} has made great progress in image generation and editing,
there are few research \cite{ricker2022towards,bohavcek2023geometric,song2023robustness} focus on training deepfake detection models with images generated by diffusion models.
In addition, directly using the images generated by diffusion models will lead to a series of problems, for example, 
making the detection model focus on unnecessary (e.g. non-facial) information,
thus causing a performance decline when tested on other face datasets.
\begin{figure}[t]
\centering
    \includegraphics[width=1.0\linewidth]{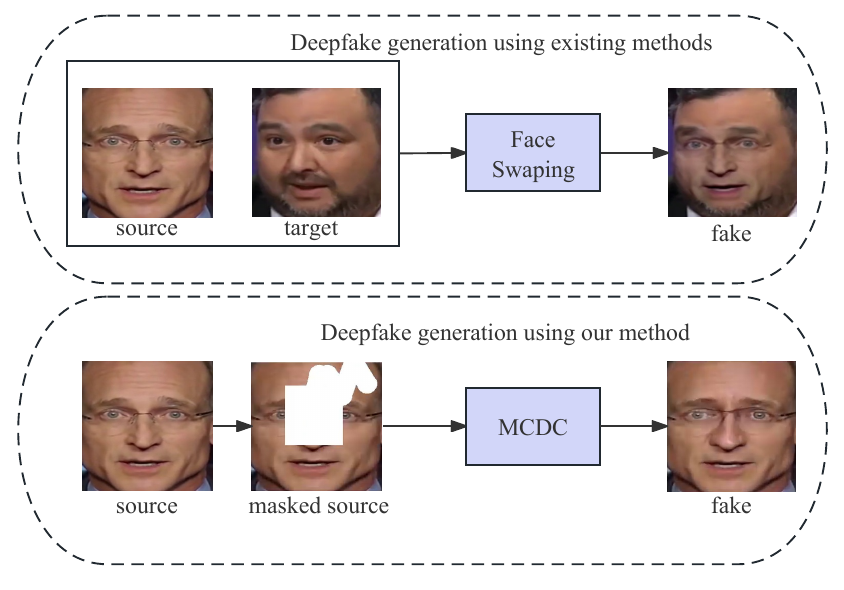}
    \caption{
        Pipeline of fake sample generation. The previous method generates samples (\textbf{top}) by swapping the two face regions. By contrast, our method generates samples (\textbf{bottom}) by complementing the face region with our MCDM. Details of MCDM are described in section \ref{section3}.
    }
    \label{fig:overall}
\end{figure}

In this paper, to address this problem, we propose a novel approach that extends the existing deepfake datasets with a diffusion model. The pipeline of our method and previous methods \cite{li2020face}, \cite{soft-cite001} are shown in Fig. \ref{fig:overall}.
In contrast to previous work that generates samples by swapping the two face regions, 
our method masks the face partially and then feeds it into our model (MCDM), outputting images that have hardly recognizable artifacts which prompts the face forgery detection model to be more general and robust.
We train the diffusion model using fixed and random shape masks as conditions. 
This preserves more original image features in the mask-outside region of the generated images.
In addition, we introduce a feature-level reconstruction loss during the training process, 
which encourages the generated images to be similar to the original counterpart not only in the pixel space 
but also in the feature semantic information space.
In summary, the main contributions of this paper can be summarized as follows:

\begin{itemize}
\item  To the best of our knowledge, we are the first to use conditional diffusion model-generated images for data augmentation in the realm of deepfake detection.
\item  We propose a mask-based approach to manipulate the original image while focusing on its intrinsic semantic features throughout the complementation process.
\item  Extensive experiments demonstrate the significance of our proposed framework in improving the deepfake detection performance. \end{itemize}

\begin{figure*}[t]
    \centering
    \includegraphics[width=0.99\linewidth]{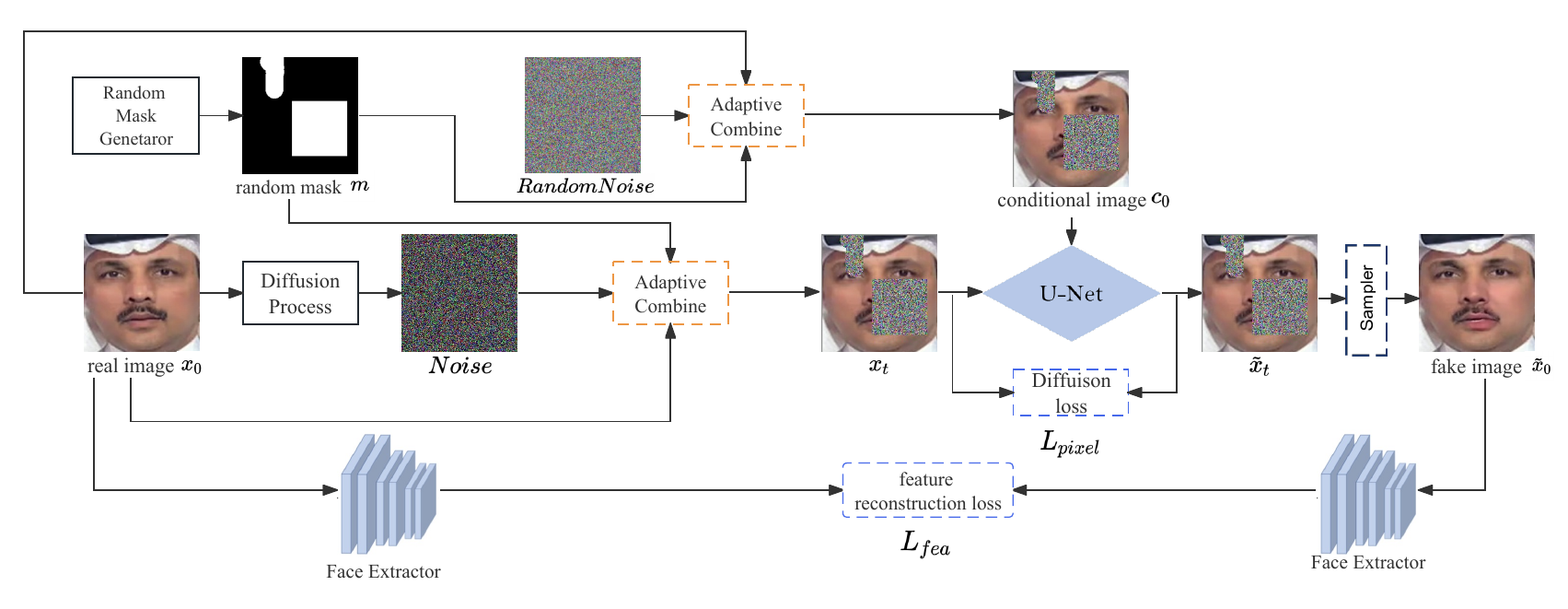}
    \caption{The overview architecture of the proposed method. The input (real) images first enter the diffusion forward process. diffusion process generates noise images from the real image using the diffusion formula. 
    The random mask $m$ generated with the random mask generator and the random noisy image are hybridized by the adaptive combine module and subsequently fed into the diffusion inverse process to get the output (fake) image. The whole system is trained by jointly minimizing the diffusion loss $L_{pixel}$ and the reconstruction loss $L_{fea}$ .}
    \label{fig:framework}
\end{figure*}

\section{Related Works}
This section provides an overview of deepfake generation, deepfake detection, and generative methods for deepfake data augmentation.

\subsection{Deepfake Generation}

Due to the advancements in deep learning, deepfake technology has become increasingly prevalent.
Korshunova et al.~\cite{korshunova2017fast} propose a convolution neural network (CNN) based approach that frames the face-swapping problem in terms of style transfer, where the goal is to render an image in the style of another one \cite{wang2023deep}. 
Vougioukas et al.~\cite{Vougioukas_Petridis_Pantic_2019} present a system for generating videos of a talking head, using a still image of a person and an audio clip containing speech. 
AttGAN \cite{he2019attgan} proposes an encoder-decoder architecture that considers the relationship between attributes and potential representations, and it generates the deepfake by changing the potential representation of the attribute.

\subsection{Deepfake Detection}
Computer vision has made significant progress as a key technology area in addressing the challenges of deepfake forgery techniques, from a binary (real or forgery) classification problem to adapting neural networks to extract discriminative features for deepfake detection automatically. 
For instance, Zhao et al. \cite{zhao2021learning} propose to detect deepfakes generated by face manipulation based on one of their fundamental features. 
Rossler et al. \cite{rossler2019faceforensics++} propose a deepfake dataset (FaceForensics++) for training deepfake detection models. 
They use a convolutional neural network-based approach for detection by learning associations between facial features and signs of manipulation. 
Zhou et al. \cite{Zhou_Han_Morariu_Davis_2017} present a two-stream tampered face detection technique, where one stream detects low-level inconsistencies between image patches and another stream explicitly detects tampered faces.
In addition, robust learning methods have been developed to address the issue of imbalanced learning in deepfake detection \cite{pu2022learning,guo2022robust,hu2023rank}.
Although several deepfake detection works\cite{shao2022detecting,wang2022deepfake,fei2022learning,cao2022end, fan2023synthesizing, zhang2023x, yang2023improving, fan2023attacking, ju2024improving, guo2022open, guo2022eyes, hu2021exposing} have been proposed and show exciting performance in intra-dataset scenarios where training and testing images are generated with the same deepfake technology, 
their performance tends to decline when confronted with inter-dataset scenarios where the tested faces are from unknown domains.

\subsection{Methods toward Deepfake Data Augmentation}
To improve the generalization ability of deepfake detection models, deepfake data augmentation methods have received increasing attention in recent years.
Li et al. \cite{li2020face} present a landmark-based generation method to synthesize new images. 
% Instead of simply using a landmark-oriented synthesis strategy, 
Chen et al. \cite{chen2022self} provide an adversarial training strategy to synthesize challenging forgeries to the current detection model dynamically.
Shiohara et al. \cite{shiohara2022detecting}  present a localization architecture that generates images by blending pseudo source and target images from single pristine images. Hu et al.\cite{hu2023mover} utilize a masked autoencoder to reconstruct missing areas based on the remaining facial parts.
Differing from these methods, we employed a masked conditional diffusion model for deepfake data augmentation, 
which enhances the authenticity of generated images from pixel and feature levels.

\section{Our Method}
\label{section3}
The architecture of our proposed method is shown in Fig.~\ref{fig:framework}.
Given any input image $x_{0}\in R^{3\times H\times W}$ ($H, W$ represents the height and width, respectively), and 
a mask image $m \in R^{3\times H\times W}$ from the random mask generator.
The input image $x_{0}$ is progressively noised to end up with a completely noisy one $Noise$.
We input $Noise$ and $x_{0}$ and $m$ into the adaptive combine module to get $x_{t}$.

Meanwhile, we create a conditional image $c_{0}$. Specifically, we first generate a random noise $RandomNoise$, and after that, we feed $RandomNoise$ along with the mask $m$ and $x_{0}$ into the adaptive combine module to get $c_{0}$. 
The $x_{t}$ and $c_{0}$ will be fed into the model to obtain a noise estimate that stepwise complements the face part from the noise.
With the randomly generated masks and resulting inpainted face images, we can enhance the diversity and scale of the deepfake dataset.

\subsection{Random Mask generating}

This block receives the size of the image that will be masked and outputs a binarised masked image of the same size. The generating algorithm is shown in Algorithm \ref{algorithm1}. Specifically, we initialize an image with a value of 0, randomly select a coordinate point $(X_1, Y_1)$ in the image, and draw a circle with $(X_1, Y_1)$ as the circle's center. Then, based on the randomly obtained angle and another coordinate point $(X_2, Y_2)$, we draw a straight line from the first coordinate point to the second coordinate point. These steps are iteratively applied until the random point reaches the boundary of the image. Finally, we add a random square to the mask image.

\subsection{Mask-Conditioned Diffusion Modeling}

Although the original diffusion model \cite{Nichol_Dhariwal_2021} (generating images directly without masks) can generate high-quality images with spatial continuity, 
it is not suitable for deepfake image augmentation because the key to generating deepfake face images is to alter the real facial regions.
Moreover, a primitive mask strategy with a high masking rate will cause the model to recover its original appearance with unnecessary details. 
Therefore, we propose a mask strategy that utilizes a relatively low masking rate to ensure the accuracy of reconstructed images meanwhile 
emphasizing the diversity of generated content.

Specifically, we adopt the U-net architecture, which is widely used in diffusion modeling inversion processes, with the masked original image as a condition. 
Given the original image ${x}_{0}$,  %$\mathbf{x}_{0}$, ysm
we send it to the adaptive combine module with a random noise image $RandomNoise$ and random mask $m$ from the random mask generator. Then, we obtain the conditional image ${c}_{0}$. The adaptive combine module can be formulated as follows:
\begin{align}
c_{0}=\left(1-m\right) \odot x_{0}+m \odot RandomNoise
\end{align}
Applying an optimized Markovian noise addition process to image ${x}_{0}$, we obtain an image $x_{t}$ with the help of the adaptive combine module. These processes can be expressed as follows: 
\begin{equation*}
\begin{aligned}
    {Noise}=\sqrt{\bar{\alpha}_{t}} {x}_{0}+\sqrt{1-\bar{\alpha}_{t}} \boldsymbol{\epsilon} \\
   {x_{t}=\left(1-m\right) \odot x_{0}+m \odot Noise} 
\end{aligned}
\end{equation*}
where $\boldsymbol{\epsilon}  \sim  \mathcal{N}(0,\mathrm{I})$ means Standard Gaussian noise, $\overline{\alpha}_{t}$ is a parameter used for balancing the noise intensity and the image $x_{0}$ at any time step $t$. 
Then, $ {x}_{t}$ is fed into the encoder $\mathcal{E}$ of the U-Net network combined with the condition ${c}_{0}$. The decoder $\mathcal{D}$ outputs an estimation of the noise part of image ${x}_{t}$, presented as ${\widetilde{x} _{t}}$. We formulated the decoding process as follows:
\begin{align}
    \boldsymbol{\epsilon}_{\theta}\left( {x}_{t}, t,  {c}_{0}\right)=\mathcal{D}\left(\mathcal{E}\left(concat({x}_{t},{c}_{0})\right), t\right)  .
   \end{align}
where $concat(.,.)$ means stacking the conditional and noise image along the batch dimension. 
${\epsilon}_{\theta}$ indicates the noise prediction at the current time step. Removing the noise prediction ${\epsilon}_{\theta}$ from the $ {x}_{t}$ iteratively will lead to the final result of ${\widetilde{x} _{0}}$.

\begin{algorithm}[t]
\footnotesize
\caption{Algorithm for generating a random shape mask}
\label{algorithm1}
\KwIn{\\
Height of the original image for masking, labeled as $imageHeigh$.\\
Width of the original image for masking, labeled as $imageWidth$.\\
Line width for connecting brushes, labeled as $length$.\\
Angle range of brush movement, labeled as $maxangle$.}
 
\KwOut{$\mathbf{Mask}$}
{Set $X_1 = random(0, imageHeight)$, 

$Y_1 = random(0, imageWidth)$}\;

{$\mathbf{Mask} = zeros(imageHeight, imageWidth)$,

$angle = random(0,maxAngle)$}\;

\While{$X_1 \leq imageHeight $ and $Y_1 \leq imageWidth$}
{Draw a circle at $(X_1, Y_1)$ with radius as half of length on $\mathbf{Mask}$;

$angle = random(0,maxAngle)$;

$X_2 = X_1 + length * sin(angle)$\;

$Y_2 = Y_1 + length * cos(angle)$\;

Draw line from $(X_1, Y_1)$ to $(X_2, Y_2)$  on $\mathbf{Mask}$;

$X_1 = X_2$;

$Y_1 = Y_2$;
 
 }

\end{algorithm}

\begin{figure*}[t]
    \centering
    \includegraphics[width=0.98\linewidth]{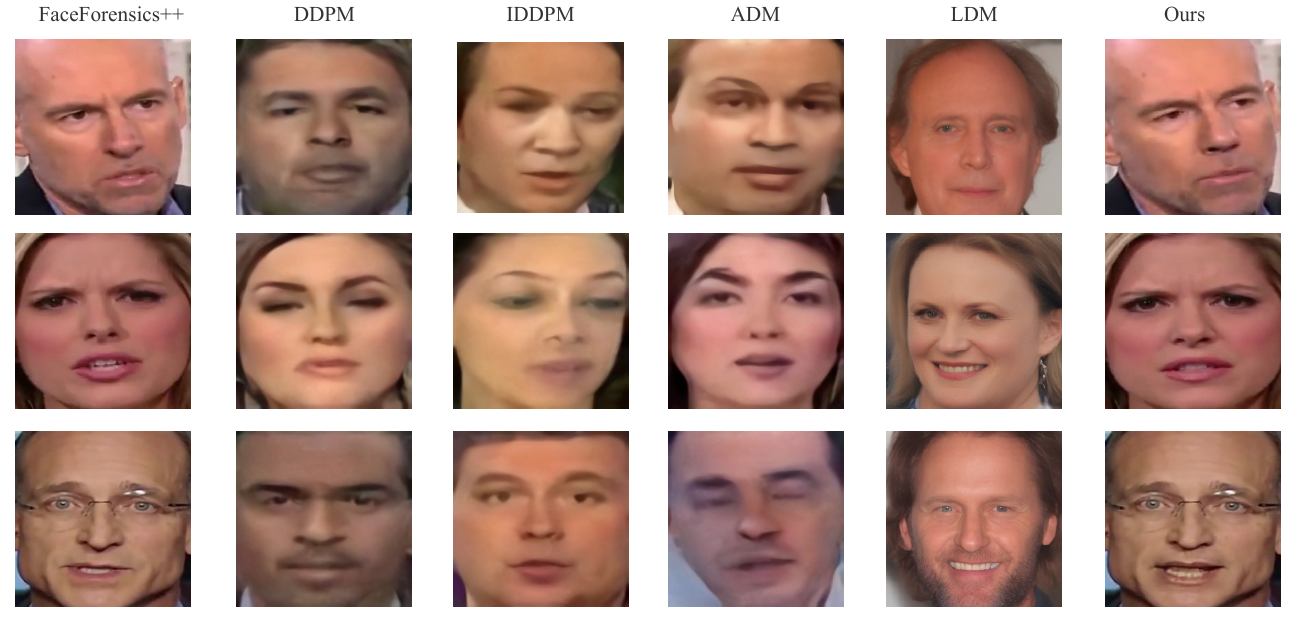}
    \caption{Visualization of deepfake images generated by different methods.
    From left to right are the real images sampled from the FF++ dataset, 
    and corresponding deepfake images generated by DDPM \cite{ho2020denoising}, IDDPM \cite{Nichol_Dhariwal_2021}, ADM \cite{dhariwal2021diffusion}, LDM \cite{Rombach_Blattmann_Lorenz_Esser_Ommer_2022}, and our proposed method, respectively. 
    % The last line indicates the mask used by the method. 
    }
    \label{fig:quanlity}
\end{figure*}

\subsection{Overall Loss}
The recovered image at each time step $t$ is expected to be as close to the original image as possible, 
and the finally generated image $\Tilde{x}_0$ is expected to be indistinguishable by deepfake detection models.    
To this end, we proposed a loss function as follows:
\begin{align*}
\mathcal{L}& = \lambda_1 \mathcal{L}_{pixel}+ \lambda_2 \mathcal{L}_{fea}\\
\mathcal{L}_{pixel}&=\mathbb{E}_{t, {x}_{t}, {c}_{0}, \boldsymbol{\epsilon} \sim \mathcal{N}(\mathbf{0}, \mathbf{I})}\left[\left\|\boldsymbol{\epsilon}-\boldsymbol{\epsilon}_{\theta}\left({x}_{t}, t, {c}_{0}\right)\right\|^{2}\right] .\\
\mathcal{L}_{fea}&=D_{cos}(G({x_0}), G(\Tilde{x}_0))
\end{align*}
where $G(\cdot)$ is the feature-extraction process of input image, and $G$ can be any pre-trained feature extracting deep neural networks 
(such as EfficientNet-B4\cite{tan2019efficientnet}, ResNet50\cite{He_Zhang_Ren_Sun_2016}, etc.). $D_{cos}(\cdot, \cdot)$ is the cosine distance between the two inputs. $\lambda_1$ and $\lambda_2$ are the balancing weights for these terms. By default, we set $\lambda_1$ = 1, $\lambda_2$ = 0.001 in all our experiments.

\begin{table}[t]
\centering

\caption{FID of images generated by different methods.}
\begin{tabular}{clll}
\hline

Methods         &  &   FID \textdownarrow          \\ \hline
DDPM \cite{ho2020denoising}     &        & 6.36         \\
IDDPM \cite{Nichol_Dhariwal_2021}   &   & 4.24         \\
ADM \cite{dhariwal2021diffusion}    &    & \textbf{1.90} \\
LDM \cite{Rombach_Blattmann_Lorenz_Esser_Ommer_2022}    &    & 2.95         \\
Ours           &    & 2.04         \\ \hline
\end{tabular}
\label{tab:table1}
\end{table}

\section{Experiment}

\subsection{Settings}
\subsubsection{Datasets} 

The FaceForensics++ \cite{rossler2019faceforensics++} (FF++) dataset is adopted in our experiments, which contains 1000 original videos from YouTube and corresponding fake videos generated by four deepfake methods, including Deepfakes \cite{Taverna_2019} (DF), Face2Face \cite{Thies_Zollhofer_Stamminger_Theobalt_Niessner_2016} (F2F), FaceSwap \cite{soft-cite001} (FS) and NeuralTextures \cite{Thies_Zollhöfer_Nießner_2019} (NT). 
Additionally, to evaluate the generalization ability of our method, we use the recently proposed Celeb-DFv2 \cite{Li_Yang_Sun_Qi_Lyu_2020} (CDF) and DeepFakeDetection \cite{Rössler_Cozzolino_Verdoliva_Riess_Thies_Nießner_2018} (DFD) deepfake dataset for the cross-dataset test. 
Celeb-DFv2 contains 5639 high-quality deepfake videos generated using an improved synthesis process. 
DFD contains 363 authentic videos from YouTube and 3068 fabricated videos.

Following the official protocols in \cite{rossler2019faceforensics++}, 720/140/140 videos from the FF++ dataset are used for training/validation/testing, respectively.
Specifically, to maintain a balance between real and fake data, four frames per fake video and eight frames per real video 
are randomly sampled from the training set of the FF++ dataset, denoted as ``baseline" in subsequent experiments. 
Additionally, 32 frames per video from the FF++, CDF, and DFD datasets are randomly sampled for testing.

\subsubsection{Evaluation Metrics}
In our experiments, Fréchet Inception Distance (FID) \cite{heusel2017gans} is adopted 
as the evaluation metric of deepfake generation.
In addition, the commonly used area under the receiver operating characteristic curve (AUC) is adopted 
to evaluate the deepfake detection performance.

\begin{table}[t]
\centering
\caption{Performance evaluation of deepfake models trained with different datasets in both intra-dataset and cross-dataset scenarios.}

\begin{tabular}{lccc}
\hline
\multicolumn{1}{c}{\multirow{3}{*}{Method}} & \multicolumn{3}{c}{Testing set AUC(\%)\textuparrow}               \\ \cline{2-4} 
\multicolumn{1}{c}{}                        & Intra-dataset    & \multicolumn{2}{c}{Cross-dataset}  \\ \cline{2-4}
\multicolumn{1}{c}{}                        & FF++             & CDF             & DFD             \\ \hline
baseline                                    & 98.35            & 73.12           & 87.80           \\
baseline+subset2                            & 98.88            & 73.92           & 89.73           \\
baseline+ADM\cite{dhariwal2021diffusion}$^*$                                & 98.83            & 73.35           & 88.75           \\
 Baseline+Ours$^*$                                        & \textbf{99.31}            & \textbf{78.02}           & \textbf{90.30}           \\ \hline
\end{tabular}
\label{tab:cross-data-test}
\end{table}

\subsubsection{Baseline Methods}

In the experiment of image generation, we compare with the following state-of-the-art baselines: DDPM \cite{ho2020denoising}, IDDPM \cite{Nichol_Dhariwal_2021}, ADM \cite{dhariwal2021diffusion}, and LDM \cite{Rombach_Blattmann_Lorenz_Esser_Ommer_2022}.

In the experiment of data enhancement effectiveness evaluation, we compare with the following baseline models: 
(i) Detection model trained using the ``baseline" subset of the FF++ dataset.
(ii) Detection model trained using the ``baseline" subset and another subset called ``subset2" 
which contain 5760 fake images randomly sampled from the FF++ dataset (2 frames per fake video from the training set of the FF++ dataset). 
(iii) Detection model trained using the ``baseline" subset and an additional subset generated with the ADM method. This additional subset is generated to have the same scale as ``subset2".

\subsubsection{Implementation Details}
For each video frame, face crops are detected by using the MTCNN \cite{Thies_Zollhofer_Stamminger_Theobalt_Niessner_2016} algorithm. All face crops are resized to 256 × 256.
Palette \cite{saharia2022palette} pre-trained on the CelebA-HQ \cite{Karras_Aila_Laine_Lehtinen_2018} dataset is adopted as our backbone for deepfake image generation, 
and EfficientNet-B4 pre-trained on the ImageNet \cite{Deng_Dong_Socher_Li_KaiLi_LiFei-Fei_2009} dataset is adopted 
as our feature extraction model for deepfake detection. 

We train the image generation model for 50 epochs with Adam \cite{Kingma_Ba_2014} as the optimizer, a batch size of 8, and a learning rate of 5e-5.

\begin{figure}[t]
\centering
    \includegraphics[width=0.98\linewidth]{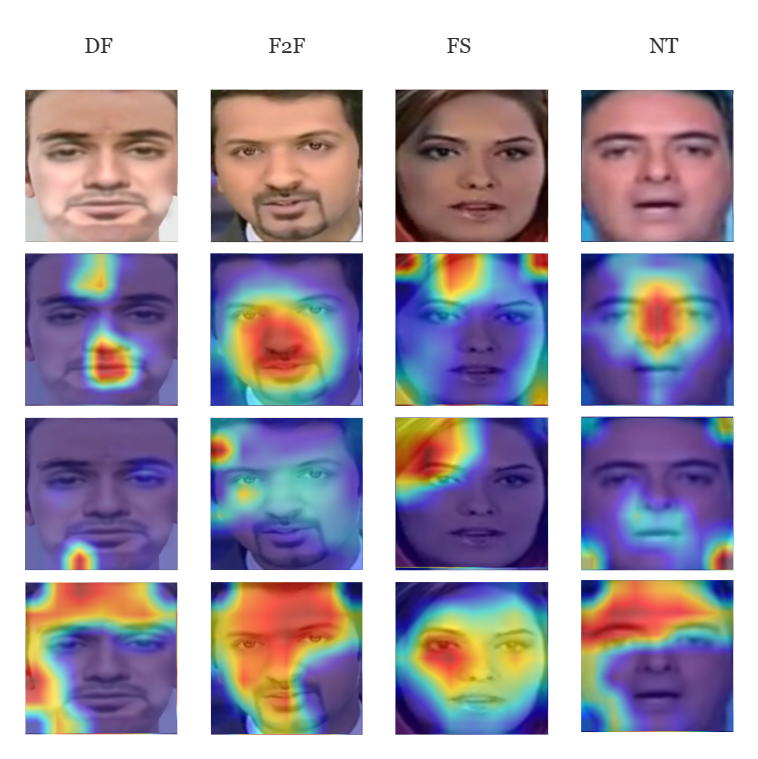} 
    \caption{Grad-CAM \cite{Selvaraju_Cogswell_Das_Vedantam_Parikh_Batra_2020} visualization of different models. 
    From top to bottom are the input images, the Grad-CAM of the model trained on ``baseline" and ``subset2", 
    the Grad-CAM of the model trained on  ``baseline" and an additional subset, having the same scale as ``subset2",
    generated with the ADM method, and the Grad-CAM of the model trained on ``baseline" and an additional subset, having the same scale as ``subset2", generated by our proposed method. 
   } % 
    \label{fig:saliency}
\end{figure}

\subsection{Results}
\subsubsection{Image Generation Comparisons}

Table \ref{tab:table1} presents the comparison results of our proposed method with other state-of-the-art diffusion models. 
It can be seen that our method performs competitively with the best method ADM in terms of FID and outperforms all the other diffusion models by a large margin.
Figure \ref{fig:quanlity} further shows the deepfake images generated by diffusion models mentioned in Table \ref{tab:table1}.
We can see that all generation methods except ours have consistency problems such as facial distortion, serious artifacts, and blurred backgrounds, while our framework can generate high-quality deepfake images with attention to facial consistency.
These results demonstrate the advantages of our method in terms of the quality of the generated images.

\subsubsection{Intra-Dataset Deepfake Detection Evaluation}
We evaluated our proposed method with the intra-dataset situation. Table \ref{tab:cross-data-test} shows the intra-dataset evaluation results and comparison with the baselines, our method outperforms FF++ and ADM, achieving the best 99.31\%.
\subsubsection{Cross-Dataset Deepfake Detection Evaluation}

To validate the effectiveness of deepfake images generated by our proposed method, cross-dataset deepfake detection evaluation is conducted in this section. Table \ref{tab:cross-data-test} shows the experimental results.
We can see that,
all three models trained with more deepfake images exhibit higher AUC performance on CDF and DFD (unseen during training) test datasets than the one trained with only the ``baseline" subset. 
Specifically, our method exhibits the best performance, surpassing 
the other two models, one trained on the ``baseline" and ``subset2" and the other trained on ``baseline" and images generated by ADM technology with the same size of subset2, by a margin of 5.55\%  and 6.37\%  on the CDF test dataset, respectively. 
On the DFD test dataset, the performance gain is 0.64\% and 1.75\%, respectively.
These results emphasize the advantages of our method in enhancing the performance of deepfake detection.

\begin{table}[t]
\centering
\caption{Ablation study  of the conditional mask and feature-level reconstruction loss $L_{fea}$ on  deepfake detection.}
\begin{tabular}{ccclcl}
\hline
\multicolumn{2}{c}{Methods}                         & \multicolumn{4}{c}{Testing set AUC (\%)}\textuparrow                                    \\ \cline{3-6}
\multicolumn{1}{l}{mask} & \multicolumn{1}{l}{$\mathcal{L}_{fea}$} & \multicolumn{2}{c}{CDF}            & \multicolumn{2}{c}{DFD}              \\ \hline
\XSolidBrush                        & \XSolidBrush                        & \multicolumn{2}{c}{70.67}          & \multicolumn{2}{c}{86.27}            \\
\Checkmark                        & \XSolidBrush                        & \multicolumn{2}{c}{74.58}          & \multicolumn{2}{c}{89.36}            \\
\XSolidBrush                        & \Checkmark                        & \multicolumn{2}{c}{73.96}          & \multicolumn{2}{c}{89.13}            \\
\Checkmark                        & \Checkmark                        & \multicolumn{2}{c}{\textbf{78.02}} & \multicolumn{2}{c}{{\textbf{90.30}}} \\ \hline
\end{tabular}
\label{tab:ablation}
\end{table}

\subsubsection{Visualization}
Fig. \ref{fig:saliency} further illustrates the deepfake locations of different models on deepfake images (from the FF++ dataset) generated using DF, F2F, FS, and NT technologies. It can be observed that our method encourages the model to pay more attention to the forged facial borders compared to other methods. Consequently, we can precisely locate the boundaries of the forged face, leading to an improvement in detection performance.

\subsection{Ablation Study}
In this section, we perform ablation studies to verify the effectiveness of the proposed mask condition and feature reconstruction loss. All the models are trained on the same dataset (subset1 of FF++ and the corresponding generated data) and tested on the CDF and DFD.

\subsubsection{Effect of Mask-conditioning}
In the training phase of the diffusion model, we utilized randomly masked images as input conditions. Notably, the artifacts intentionally introduced into our model influence the detection model learned representations. To assess the efficacy of Mask-conditioning, we conducted ablation experiments on the generation process. Specifically, we trained our model without incorporating Mask-conditioning. The results of this experimental comparison are presented in Table \ref{tab:ablation}. By comparing the first and second lines, we observe that the proposed mask-conditioning module brings 5.5\% and 4.7\% AUC gains on the CDF and DFD datasets, respectively.

\subsubsection{Effect of feature reconstruction loss}
As shown in the second and last rows of Table \ref{tab:ablation}, incorporating the reconstruction loss module can improve the generality of deepfake detection model on cross-dataset, enhancing the AUC by 4.6\% and 1\% on the CDF and DFD dataset, respectively. 
The results validate the benefit of using reconstruction loss during training and also suggest that the improved image quality plays an important role in the success of the deepfake detection model.

\section{Conclusion}

In this paper, we present a novel approach to augment face forgery data. Specifically, we introduce a diffusion model tailored for image restoration, enabling the seamless reconstruction of regions occluded by random masks, thereby generating diverse deepfake images. Our methodology incorporates pixel-level and feature-level reconstruction losses to enhance the fidelity of image reconstruction. Experimental results across multiple challenging datasets demonstrate that our proposed method exhibits superior generalization capability compared to existing diffusion models. In the future, we plan to integrate this conditional diffusion model-based data augmentation method with a deepfake detection network in an end-to-end manner, aiming to further enrich the diversity of generated images and enhance the performance of deepfake detection.

\bibliographystyle{IEEEtran}

\bibliography{references}

\vspace{12pt}

\end{document}